\documentclass[10pt,twocolumn,letterpaper]{article}

\usepackage{iccv}
\usepackage{times}
\usepackage{epsfig}
\usepackage{graphicx}
\usepackage{amsmath}
\usepackage{amssymb}

\usepackage[breaklinks=true,bookmarks=false]{hyperref}

\iccvfinalcopy 

\def\iccvPaperID{6296} 

\ificcvfinal\fi

\usepackage[usenames,dvipsnames]{xcolor}

\usepackage{iccv}
\usepackage{times}
\usepackage{epsfig}
\usepackage{graphicx}
\usepackage{amsmath}
\usepackage{amssymb}

\usepackage{algorithmic}
\usepackage{algorithm}
\usepackage{rotating}
\usepackage{diagbox}
\usepackage{balance}

\usepackage[utf8]{inputenc} 
\usepackage[T1]{fontenc}    
\usepackage{url}            
\usepackage{booktabs}       
\usepackage{amsfonts}       
\usepackage{amsmath}
\usepackage{nicefrac}       
\usepackage{microtype}      
\usepackage{graphicx}       

\usepackage{paralist}
\usepackage{xspace}
\usepackage{comment}
\usepackage{subfig}

\newcommand{\myvector}[1]{\boldsymbol{#1}}

\newcommand{\modelname}{SPL}

\newcommand{\modelnamelayer}{SP-layer}

\newcommand{\poset}[1]{\myvector{x}_{#1}}
\newcommand{\pose}{\poset{t}}
\newcommand{\posehat}{\myvector{\hat{x}}_t}
\newcommand{\posepos}{\myvector{p}_t}
\newcommand{\poseposhat}{\myvector{\hat{p}}_t}

\newcommand{\joint}[1]{\myvector{x}_t^{(#1)}}
\newcommand{\jointhat}[1]{\myvector{\hat{x}}_t^{(#1)}}
\newcommand{\jointpos}[1]{\myvector{p}_t^{(#1)}}
\newcommand{\jointposhat}[1]{\myvector{\hat{p}}_t^{(#1)}}

\newcommand{\figref}[1]{Fig.~\ref{#1}}
\newcommand{\eqnref}[1]{Eq.~\ref{#1}}
\newcommand{\tabref}[1]{Tab.~\ref{#1}}
\newcommand{\secref}[1]{Sec.~\ref{#1}}
\newcommand{\norm}[1]{\left\lVert#1\right\rVert}
\newcommand{\real}[0]{\Bbb R}

\begin{document}

\newif\ifsupplementary
\supplementarytrue
\newif\ifappendixsupplementary
\appendixsupplementarytrue

\ifappendixsupplementary
    \title{Structured Prediction Helps 3D Human Motion Modelling}
    
    \author{Emre Aksan\textsuperscript{*} \quad \quad Manuel Kaufmann\textsuperscript{*} \quad \quad Otmar Hilliges\\
    Department of Computer Science, ETH Z\"urich\\
    {\tt\small \{firstname.lastname\}@inf.ethz.ch}}
    
    \twocolumn[{
        \renewcommand\twocolumn[1][]{#1}%
        \vspace{-3em}
        \maketitle
        \vspace{-3em}
        \begin{center}
    	\centering
    	\includegraphics[width=1.0\textwidth]{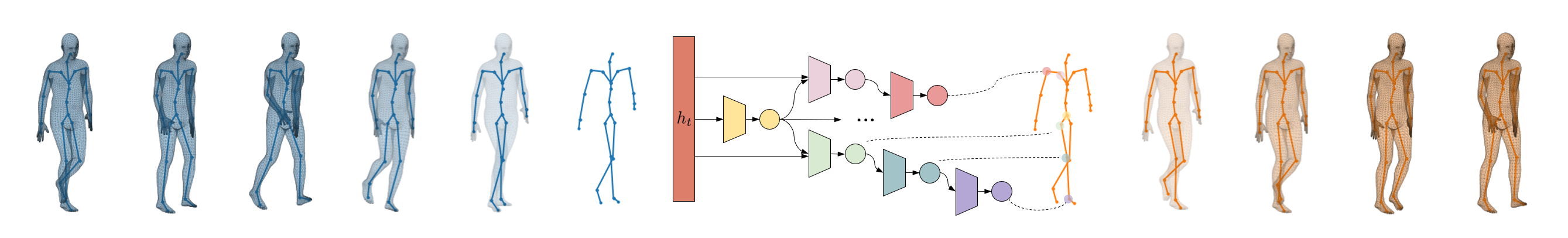}
    	\vspace{-2.5em}
    	\captionof{figure}{
    	We introduce a structured prediction layer (\modelname) to the task of 3D human motion modelling.
    	The \modelnamelayer~explicitly decomposes the pose into individual joints and can be interfaced with a variety of baseline architectures.
    	We show that on H3.6M and a recent, much larger dataset, AMASS, a variety of baseline models benefit when augmented with an \modelnamelayer.}
    	\label{fig:teaser}
        \end{center}
    }]
    \ificcvfinal\thispagestyle{empty}\fi
    
    \newcommand{\customfootnotetext}[2]{{
    \renewcommand{\thefootnote}{#1}
    \footnotetext[0]{#2}}}
    \ificcvfinal\customfootnotetext{*}{The first two authors contributed equally.}\fi

    \begin{abstract}
        Human motion prediction is a challenging and important task in many computer vision application domains. Existing work only implicitly models the spatial structure of the human skeleton. In this paper, we propose a novel approach that decomposes the prediction into individual joints by means of a structured prediction layer that explicitly models the joint dependencies. This is implemented via a hierarchy of small-sized neural networks connected analogously to the kinematic chains in the human body as well as a joint-wise decomposition in the loss function. The proposed layer is agnostic to the underlying network and can be used with existing architectures for motion modelling. 
        Prior work typically leverages the H3.6M dataset. We show that some state-of-the-art techniques do not perform well when trained and tested on AMASS, a recently released dataset 14 times the size of H3.6M.
        Our experiments indicate that the proposed layer increases the performance of motion forecasting irrespective of the base network, joint-angle representation, and prediction horizon. 
        We furthermore show that the layer also improves motion predictions qualitatively. We make code and models publicly available at \small{\url{https://ait.ethz.ch/projects/2019/spl}}.
    \end{abstract}
    
    \section{Introduction}
Modelling of human motion over time has a number of applications in activity recognition, human computer interaction, human detection and tracking, and image-based pose estimation in the context of robotics or self-driving vehicles. Humans have the ability to forecast the sequence of poses over short-term horizons with high accuracy and can imagine probable motion over arbitrary time scales. Despite recent progress in data-driven modelling of human motion \cite{Fragkiadaki2015ERD,Ghosh2017,Jain2016, Martinez2017Motion,Pavllo2018, Wang2018Adversarial}, this task remains difficult for machines. 

The difficulty of the task is manifold. First, human motion is highly dynamic, non-linear and over time becomes a stochastic sequential process with a high degree of inherent uncertainty. Humans leverage strong structural and temporal priors about continuity and regularity in natural motion. However, these are hard to model algorithmically due to 
\begin{inparaenum}[i)]
\item the inter-dependencies between joints and 
\item the influence of high-level activities on the motion sequences (e.g., transition from walking to jumping). 
\end{inparaenum}
In fact many recent approaches forgo explicit modelling of human motion \cite{Jain2016} in favor of pure data-driven models \cite{Ghosh2017,Martinez2017Motion,Pavllo2018}. 

Initial Deep Learning-based motion modelling approaches have focused on recurrent neural networks (RNNs) \cite{Ghosh2017, Fragkiadaki2015ERD,Jain2016}, using curriculum learning schemes to increase robustness to temporal drift. Martinez \etal{} \cite{Martinez2017Motion}
have shown that a simple running-average provides a surprisingly difficult to beat baseline in terms of Euler angle error. Following this, sequence-to-sequence models trained in an auto-regressive fashion have been proposed \cite{Martinez2017Motion}, sometimes using adversarial training to address the drift problem in long-term predictions \cite{Wang2018Adversarial}. Pavllo \etal \cite{Pavllo2018} study the impact of joint angle representation and show that a quaternion-based parameterization improves short-term predictions. 

However, it has been observed that quantitative performance does not always translate to qualitatively meaningful predictions \cite{Martinez2017Motion,Pavllo2018}. Furthermore, the H3.6M benchmark is becoming saturated, limiting progress. This leads to the two main research questions studied in this work:
\begin{inparaenum}[i)]
\item How to measure accuracy of pose predictions in a meaningful way such that low errors corresponds to good qualitative results and how to improve this performance? 
\item How to exploit spatial structure of the human skeleton for better predictions?
\end{inparaenum}

With respect to i) we note that much of the literature relies on the H3.6M \cite{h36m} dataset and an Euler angle based metric as performance measure, evaluated on a limited number of test sequences. While enabling initial exploration of the task, the dataset is limited in size (roughly 3 hours from 210 sequences) and in diversity of activities and poses, which contributes to a saturation effect in terms of performance. In this paper we show that existing techniques do not scale well when trained on larger and more diverse datasets. To this end, we leverage the recently released AMASS dataset \cite{AMASS2019}, itself consisting of multiple smaller motion datasets, offering many more samples (14x over H3.6M) and a wider range of activities. To further unpack the performance of motion modelling techniques, we introduce several evaluation metrics to the task of human motion prediction.

Our main technical contribution is a novel structured prediction layer (\modelname) that addresses our second research question. We leverage the compositional structure of the human skeleton by explicitly decomposing the pose into individual joints. The \modelnamelayer~models the structure of the human skeleton and hence the spatial dependencies between joints. This is achieved via a hierarchy of small-sized neural networks that are connected analogously to the kinematic chains of the human skeleton. Each node in the graph receives information about the parent node's prediction and thus information is propagated along the kinematic chain. We furthermore introduce a joint-wise decomposition of the loss function as part of \modelname. The proposed layer is agnostic to the underlying network and can be used in combination with most previously proposed architectures. 

We show experimentally that introducing this layer to existing approaches improves performance of the respective method. The impact is most pronounced on the larger and more challenging AMASS dataset.
This indicates that our approach is indeed a step towards successfully exploiting spatial priors in human motion modelling and in turn allows recurrent models to capture temporal coherency more effectively.
We thoroughly evaluate the \modelnamelayer~on H3.6M and AMASS. On AMASS, for any base model, any metric, and any input representation, it is beneficial to use the \modelnamelayer. 
Furthermore, even simple architectures that are outperformed by a zero-velocity baseline \cite{Martinez2017Motion} perform competitive if paired with the \modelnamelayer.  

In summary, we contribute:
\begin{inparaenum}[i)]
    \item An in-depth analysis of state-of-the-art motion modelling methods and their evaluation.
    \item A new benchmark and evaluation protocol on the recent, much larger AMASS dataset.
    \item A novel prediction layer, incorporating structural priors.
    \item A thorough evaluation of the \modelnamelayer's impact on motion modelling in combination with several base models.
\end{inparaenum}

    \section{Related Work}
We briefly review the most related literature on human motion modelling focusing on Deep Learning for brevity.

\paragraph{Deep recurrent models}
Early work makes use of specialized Deep Belief Networks for motion modelling \cite{Taylor2011}, whereas more recent works leverage recurrent architectures. For example, Fragkiadaki \etal \cite{Fragkiadaki2015ERD} propose the Encoder-Recurrent-Decoder (ERD) framework, which maps pose data into a latent space where it is propagated through time via an LSTM cell. The prediction at time step $t$ is fed back as the input for time step $t+1$. This scheme quickly leads to error accumulation and hence catastrophic drift over time. To increase robustness, Gaussian noise is added during training. While alleviating the drift problem, this training scheme is hard to fine-tune. Quantitative and qualitative evaluations are performed on the publicly available H3.6M dataset \cite{h36m}, with a joint angle data representation using the exponential map (also called angle-axis). The joint-wise Euclidean distance on the Euler angles is used as the evaluation metric. Most of the follow-up work adheres to this setting.

Inspired by \cite{Fragkiadaki2015ERD}, Du \etal \cite{Du2019Pedestrian} have recently proposed to combine a three-layer LSTM with bio-mechanical constraints encoded into the loss function for pedestrian pose and gait prediction. Like \cite{Du2019Pedestrian}, we also incorporate prior knowledge into our network design, but do so through a particular design of the output layer rather than enforcing physical constraints in the loss function.
Similar in spirit to \cite{Fragkiadaki2015ERD}, Ghosh \etal \cite{Ghosh2017} stabilize forecasting for long-term prediction horizons via application of dropouts on the input layer of a denoising autoencoder.
In this work we focus on short-term predictions, but also apply dropouts directly on the inputs to account for noisy predictions of the model at test time. Contrary to \cite{Ghosh2017}, our model can be trained end-to-end.

Martinez \etal \cite{Martinez2017Motion} employ a sequence-to-sequence architecture using a single layer of GRU cells \cite{GRU}. The model is trained auto-regressively, using its own predictions during training. A residual connection on the decoder leads to smoother and improved short-term predictions. Martinez \etal also show that simple running-average baselines are surprisingly difficult to beat in terms of the Euler angle metric. The currently best performance on H3.6M is reported by Wang \etal \cite{Wang2018Adversarial}. They also use a sequence-to-sequence approach trained with an adversarial loss to address the drift-problem and to create smooth predictions. Highlighting some of the issues with the previously used $L_2$ loss, \cite{Wang2018Adversarial} propose a more meaningful geodesic loss.

In this work we show that sequence-to-sequence models, despite good performance on H3.6M, do not fare as well on the larger, more diverse AMASS dataset. Although augmenting them with our \modelnamelayer~boosts their performance, they are outperformed by a simple RNN that uses the same \modelnamelayer. To better characterize motion modelling performance we furthermore introduce several new evaluation metrics.

\paragraph{Structured Prediction}
Jain \etal \cite{Jain2016} propose to explicitly model structural information by automatically converting an st-graph into an RNN (S-RNN). The skeleton is divided into 5 major clusters, whose interactions are then manually encoded into an st-graph. Our model is also structure-aware. However, our approach does not require a coarse subdivision of joints and does not require manual definition of st-graphs.  Moreover, our layer is agnostic to the underlying network and can be interfaced with most existing architectures.

Bütepage \etal \cite{Butepage2017RepL} propose to encode poses with a hierarchy of dense layers following the kinematic chain starting from the end-effectors (dubbed H-TE), which is similar to our \modelnamelayer. In contrast to this work, H-TE operates on the input rather than the output, and has only been demonstrated with non-recurrent networks when using 3D positions to parameterize the poses.

Structure-aware network architectures have also been used in 3D pose estimation from images \cite{Lee2018pLSTM, Sun2017, Moreno2017Distance, Li2015Structure, Tekin2016Structure}. \cite{Li2015Structure} and \cite{Tekin2016Structure} both learn a structured latent space.  \cite{Moreno2017Distance} exploit structure only implicitly by encoding the poses into distance matrices which then serve as inputs and outputs of the network. \cite{Lee2018pLSTM} and \cite{Sun2017} are closest to our work as they explicitly modify the network to account for skeletal structure, either via the loss function \cite{Sun2017}, or using a sequence of LSTM cells for each joint in the skeleton \cite{Lee2018pLSTM}. \cite{Lee2018pLSTM} introduces many new layers into the architecture and needs hyper-parameter tuning to be most effective. In contrast, our proposed \modelnamelayer~is simple to implement and train. We show that it improves performance of several baseline architectures out-of-the-box. 

\paragraph{Parameterizations}
Most work parameterizes joint angles as exponential maps relative to each joint's parent. Pavllo \etal \cite{Pavllo2018} show results competitive with the state of the art using quaternions. Their model, QuaterNet, consists of 2 layers of GRU cells and similar to \cite{Martinez2017Motion} uses a skip connection. The use of quaternions allows for integration of a differentiable forward kinematics layer, facilitating loss computation in the form of Euclidean distance of 3D joint positions. For short-term predictions, QuaterNet directly optimizes for the Euler-angle based metric as introduced by \cite{Fragkiadaki2015ERD}. We show that QuaterNet also benefits from augmentation with our \modelnamelayer, indicating that \modelname~is independent of the underlying joint angle representation.

Bütepage \etal \cite{Butepage2017RepL, Butepage2018Anticipating} and Holden \etal \cite{Holden2015} convert the data directly to 3D joint positions. These works do not use recurrent structures, which necessitates the extraction of fixed-size, temporal windows for training. \cite{Butepage2017RepL} and \cite{Holden2015} focus on learning of latent representations, which are shown to be helpful for various tasks, such as denoising, forecasting, or motion generation along a given trajectory \cite{Holden2016}. \cite{Butepage2018Anticipating} extends \cite{Butepage2017RepL} by applying a conditional variational autoencoder (VAE) to the task of online motion prediction in human-robot interactions. 
We use the positional representation of human poses to compute an informative metric of the prediction quality. However, for learning we use joint angles since they encode symmetries better and are inherently bone-length invariant. 
    \section{Method}
The goal of our work is to provide a general solution to the problem of human motion modelling. To this end we are motivated by the observation that human motion is strongly regulated by the spatial structure of the skeleton. However, integrating this structure into deep neural network architectures has so far not yielded better performance than architectures that only model temporal dependencies explicitly. In this section we outline a novel structured prediction layer (\modelname) that explicitly captures the spatial connectivity. The layer is designed to be agnostic to the underlying network. We empirically show in \secref{sec:h36m} and \ref{sec:amass} that it improves the performance of a variety of existing models irrespective of the dataset or the data representation used. 

\subsection{Problem Formulation}
A motion sample can be considered as a sequence $\myvector{X} = \{\poset{1} \dots \poset{T}\}$ where a frame $\pose \in \real^N$ at time-step $t$ denotes the $N$-dimensional body pose. $N$ depends on the number of joints in the skeleton, $K$, and the size $M$ of the per-joint representation (angle-axis, rotation matrices, quaternions, or 3D positions), i.e. $N = K \cdot M$.

Due to their temporal nature, motion sequences are often modelled with auto-regressive approaches. Such models factorize the joint probability of a motion sequence as a product of conditionals as follows:
\begin{equation}
p_{\theta}(\myvector{X}) = \prod_{t=1}^{T} p_{\theta}(\pose \mid \poset{1:t-1}) 
\label{eq:joint-probability}
\end{equation}
where the joint distribution is parameterized by $\theta$. At each time step $t$, the next pose is predicted given the past poses. 

While this auto-regressive setting explicitly models the temporal dependencies, the spatial structure is treated only implicitly. In other words, given a pose vector $\pose$, the model must predict the whole pose vector $\poset{t+1}$ at the next time step. This assumes that joints are independent from each other given a particular context (i.e., a neural representation of the past frames). However, the human body is composed of hierarchical joints and the kinematic chain introduces spatial dependencies between them.

\subsection{Structured Prediction Layer}
To address this shortcoming, we propose a novel structured prediction layer (\modelname). This is formed by decomposing the model prediction into individual joints. This decomposition is guided by the spatial prior of the human kinematic chain, depicted in \figref{fig:spl-model}. Formally, $\pose \in \real^N$ is a concatenation of $K$ joints $\joint{k} \in \real^M$:
\begin{equation*}
\pose = [\joint{hip}, \joint{spine} \dotsc~\joint{lwrist}, \joint{lhand}]
\label{eq:joint-decomposition}
\end{equation*}

\begin{figure}[t]
	\centering
	\includegraphics[width=0.99\columnwidth, trim={0pt 0pt 0pt 0pt}]{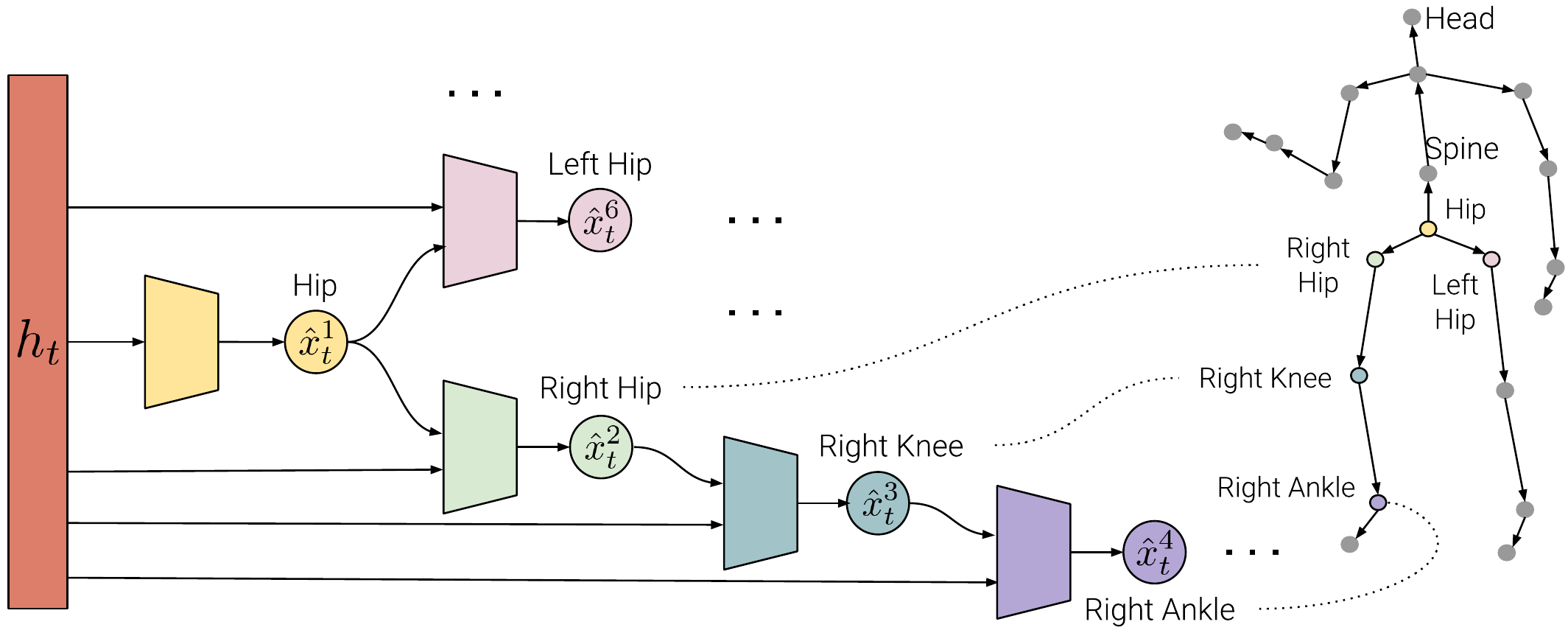}
	\caption{\textbf{SPL overview}. Given the context $\myvector{h}_t$ of past frames, joint predictions $\jointhat{k}$ are made hierarchically by following the kinematic chain defined by the underlying skeleton. Only a subset of joints is visualized for clarity. }
	\label{fig:spl-model}
\end{figure}

To interface with existing architectures, the \modelnamelayer~takes a context representation $\myvector{h}_t$ as input. Here, $\myvector{h}_t$ is assumed to summarize the motion sequence until time $t$. Without loss of generality, we assume this to be a hidden RNN state or its projection. While existing work typically leverages several dense layers to predict the $N$-dimensional pose vector $\pose$ from $\myvector{h}_t$, our \modelnamelayer~predicts each joint individually with separate smaller networks:

\begin{equation}
p_{\theta}(\pose) = \prod_{k=1}^{K} p_{\theta}(\joint{k} \mid \text{parent}(\joint{k}), \myvector{h}_{t})
\label{eq:pose-probability-structured-1}
\end{equation}
where $\text{parent}(\joint{k})$ extracts the parent of the $k$-th joint. Importantly, the full body pose $\pose$ is predicted by following the skeletal hierarchy in \figref{fig:spl-model} as follows:
\begin{equation}
p_{\theta}(\pose) = p_{\theta}(\joint{hip} \mid \myvector{h}_{t})p_{\theta}(\joint{spine} \mid \joint{hip}, \myvector{h}_{t}) \cdots 
\label{eq:pose-probability-structured-2}
\end{equation}
In this formulation each joint receives information about its own configuration and that of the immediate parent both explicitly, through the conditioning on the parent joint's prediction, and implicitly via the context $\myvector{h}_t$. The joint probability of \eqnref{eq:joint-probability} is further factorized in the spatial domain:

\begin{equation}
p_{\theta}(\myvector{X}) = \prod_{t=1}^{T} \prod_{k=1}^{K} p_{\theta}(\joint{k} \mid \text{parent}(\joint{k}), \myvector{h}_{t})
\label{eq:joint-probability-structured}
\end{equation}

The benefit of this structured prediction approach is two-fold. First, the proposed factorization allows for integration of a structural prior in the form of a hierarchical architecture where each joint is modelled by a different network. This allows the model to learn dedicated representations per joint and hence saves model capacity. Second, analogous to message passing, each parent propagates its prediction to the child joints, allowing for more precise local predictions because the joint has access to the information it depends on (i.e., the parent's prediction).

In our experiments (cf. Sec. \ref{sec:h36m} and \ref{sec:amass}) we show that this layer improves the prediction performance of a diverse set of underlying architectures across many settings and metrics. One potential reason for why this is the case can be found in the resulting network structure and its implications on network training. \figref{fig:dense_vs_spl} compares our structured approach with the traditional one-shot prediction using a dense layer. Because the per-joint decomposition leads to many small separate networks, we can think of an \modelnamelayer~as a dense layer where some connections have been set to zero explicitly by leveraging domain knowledge. This decomposition changes the gradients w.r.t. the units in the hidden layer, which are now only affected by the gradients coming from the joint hierarchy that they model. In the traditional setting, the error computed as an average over all joints can easily be distributed over all network weights in an arbitrary fashion.

\begin{figure}[t!]
\begin{center}
\includegraphics[width=0.9\linewidth, trim={10pt 160pt 370pt 0pt}]{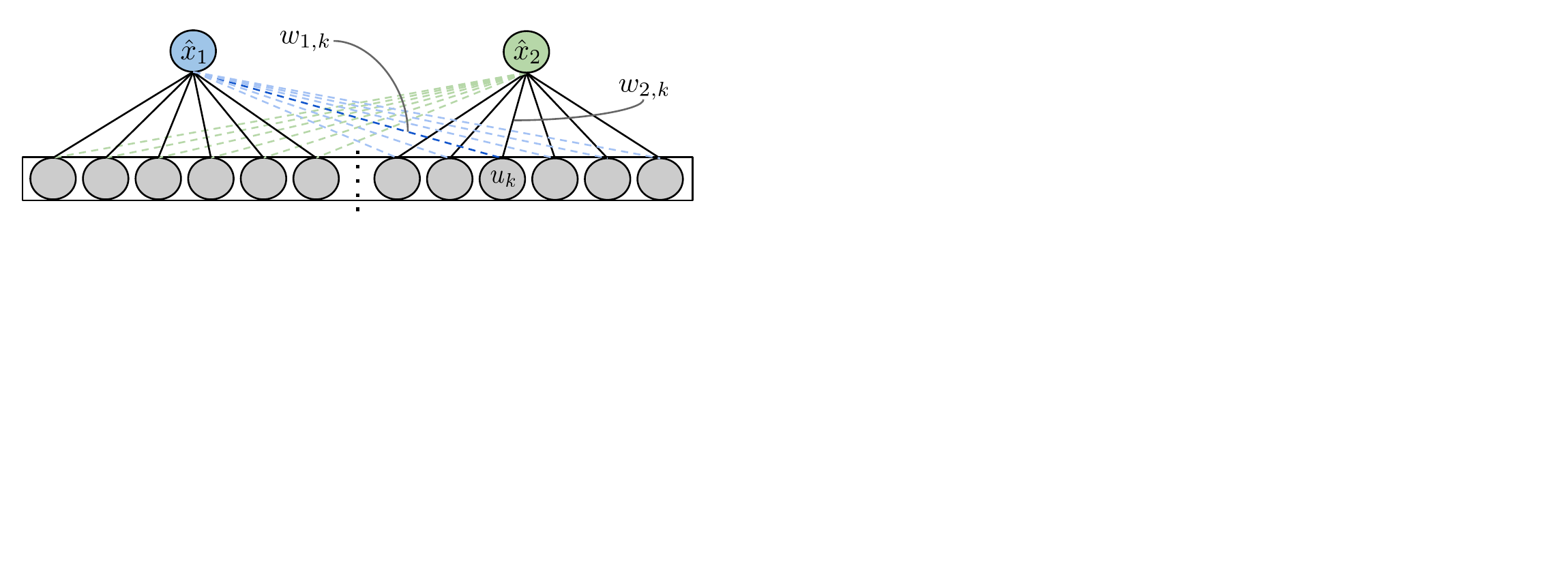}
\end{center}
   \caption{\textbf{Difference between dense and \modelnamelayer} with 2 joints. When all dashed weights are zero, a dense hidden layer is equivalent to a \modelnamelayer~that ignores the hierarchy. In a dense layer, the hidden unit $u_k$ is connected to all joints via $w_{1,k}$ and $w_{2,k}$. Hence,  the gradient $\partial L / \partial u_k$ is affected by both joints, whereas in \modelname~only $w_{2,k}$ contributes by design.
   }
\label{fig:dense_vs_spl}
\end{figure}

\subsection{Per-joint Loss}
\label{sec:per_joint_loss}
We additionally propose to perform a similar decomposition in the objective function that leads to further improvements. The training objective is often a metric in Euclidean space between ground-truth poses $\pose$ and predictions $\posehat$:
\begin{equation}
\mathcal{L}(\myvector{X}, \hat{\myvector{X}}) =  \frac{1}{T \cdot N}\sum_{t=1}^{T} f(\pose, \posehat)
\label{eq:standard_loss}
\end{equation}
where $f$ is a loss function such as an $\ell_p$ norm. The loss $f$ is calculated on the entire pose vector and averaged across the temporal and spatial domain. In our work, we use a slightly modified version that preserves joint integrity:
\begin{equation}
\mathcal{L}(\myvector{X}, \hat{\myvector{X}}) =  \sum_{t=1}^{T} \sum_{k=1}^{K} f(\joint{k}, \jointhat{k})
\label{eq:jointwise_loss}
\end{equation}
where the loss $f$ is first calculated on every joint and then summed up to calculate the loss for the entire motion sequence. In this work we use the MSE for $f$, but the formulation allows for an easy adaptation of domain-specific losses such as the \textit{geodesic distance} proposed by \cite{Wang2018Adversarial}.

    \section{Human Motion Modelling}\label{sec:hmm}
We now evaluate our \modelnamelayer~on the task of human motion modelling. We perform our experiments on two datasets and three different underlying architectures which use three different data representations. In the following we explain the datasets and models in more detail.

\subsection{Datasets}
\label{sec:dataset}
For ease of comparison to the state of the art we first report results from the H3.6M dataset. We follow the same experiment protocol used in \cite{Jain2016, Martinez2017Motion}.

Given the small size of H3.6M and the reported variance of test results \cite{Pavllo2018}, we propose to use the recently introduced AMASS dataset \cite{AMASS2019} for the motion modelling task. We downloaded the dataset from  \cite{DIP} as the data from \cite{AMASS2019} has not yet been released at the time of this writing. AMASS is composed of publicly available databases, \eg the CMU Mocap database \cite{MoCapCMU} or HumanEva \cite{sigal2010humaneva} and uses the SMPL model \cite{Loper2015SMPL} to represent motion sequences. The dataset contains $8'593$ sequences, which comprise a total of $9'084'918$ frames sampled at 60 Hz. This is roughly equivalent to $42$ hours of recording, making AMASS about 14 times bigger than H3.6M ($632'894$ frames at 50 Hz). 

We split the AMASS dataset into training, validation and test splits consisting of roughly $90\%$, $5\%$ and $5\%$ of the samples, respectively. Similar to the H3.6M protocol, the input sequences are $2$ seconds ($120$ frames) and the target sequences are $400$-ms ($24$ frames) long. The H3.6M benchmarks use a total of $120$ test samples across $15$ categories. This is a relatively small test set and it has been reported to cause high variance \cite{Pavllo2019Arxiv}. In our H3.6M experiments we use this setting to ensure fair comparison. However, on AMASS we use every frame in the test split by shifting a 2-second window over the motion sequences, which extracts $3'304$ test samples. H3.6M and AMASS model the human skeleton with $21$ and $15$ major joints, respectively. We implement separate \modelnamelayer s corresponding to the underlying skeleton. 

\begin{table*}[t!]
	\renewcommand{\arraystretch}{1.1}
	\setlength\tabcolsep{4.2pt}%
	\small{
	\begin{tabular} {l  r r r r | r r r r | r r r r | r r r r }
		& \multicolumn{4}{c}{Walking} & \multicolumn{4}{c}{Eating} & \multicolumn{4}{c}{Smoking} & \multicolumn{4}{c}{Discussion} \\       
		milliseconds & 80 & 160 & 320 & 400 & 80 & 160 & 320 & 400 & 80 & 160 & 320 & 400 & 80 & 160 & 320 & 400 \\
		\hline
		LSTM-3LR  \cite{Fragkiadaki2015ERD} & 0.77 & 1.00 & 1.29 & 1.47 & 0.89 & 1.09 & 1.35 & 1.46 & 1.34 & 1.65 & 2.04 & 2.16 & 1.88 & 2.12 & 2.25 & 2.23 \\
		SRNN \cite{Jain2016} & 0.81 & 0.94 & 1.16 & 1.30 & 0.97 & 1.14 & 1.35 & 1.46 & 1.45 & 1.68 & 1.94 & 2.08 & 1.22 & 1.49 & 1.83 & 1.93 \\
		Zero-Velocity \cite{Martinez2017Motion} & 0.39 & 0.68 & 0.99 & 1.15 & 0.27 & 0.48 & 0.73 & 0.86 & 0.26 & 0.48 & 0.97 & 0.95 & 0.31 & 0.67 & 0.94 & 1.04 \\
		AGED \cite{Wang2018Adversarial} & 0.22 & 0.36 & 0.55 & 0.67 & \textbf{0.17} & \textbf{0.28} & \textbf{0.51} & \textbf{0.64} & 0.27 & \textbf{0.43} & \textbf{0.82} & \textbf{0.84} & 0.27 & \textbf{0.56} & \textbf{0.76} & \textbf{0.83}  \\
		\hline
		Seq2seq-sampling-sup \cite{Martinez2017Motion} & 0.28 & 0.49 & 0.72 & 0.81 & 0.23 & 0.39 & 0.62 & 0.76 & 0.33 & 0.61 & 1.05 & 1.15 & 0.31 & 0.68 & 1.01 & 1.09 \\
		Seq2seq-sampling-sup-\modelname & 0.23 & 0.37 & \textbf{0.53} & \textbf{0.61} & 0.20 & 0.32 & 0.52 & 0.67 & 0.26 & 0.48 & 0.92 & 0.90 & 0.29 & 0.63 & 0.90 & 0.99 \\
		\hline
		Seq2seq-sampling \cite{Martinez2017Motion} & 0.27 & 0.47 & 0.70 & 0.78 & 0.25 & 0.43 & 0.71 & 0.87 & 0.33 & 0.61 & 1.04 & 1.19 & 0.31 & 0.69 & 1.03 & 1.12 \\
		Seq2seq-sampling-\modelname & 0.23 & 0.38 & 0.58 & 0.67 & 0.20 & 0.32 & 0.52 & 0.66 & 0.26 & 0.48 & 0.92 & 0.90 & 0.30 & 0.64 & 0.91 & 0.99\\
		\hline
		QuaterNet \cite{Pavllo2018} & \textbf{0.21} & \textbf{0.34} & 0.56 & 0.62 & 0.20 & 0.35 & 0.58 & 0.70 & \textbf{0.25} & 0.47 & 0.93 & 0.90 & \textbf{0.26} & 0.60 & 0.85 & 0.93 \\
		QuaterNet-\modelname & 0.22 & 0.35 & 0.54 & \textbf{0.61} & 0.20 & 0.33 & 0.55 & 0.68 & \textbf{0.25} & 0.47 & 0.91 & 0.88 & \textbf{0.26} & 0.59 & 0.84 & 0.91 \\
		\hline
		RNN & 0.30 & 0.48 & 0.78 & 0.89 & 0.23 & 0.36 & 0.57 & 0.72 & 0.26 & 0.49 & 0.97 & 0.95 & 0.31 & 0.67 & 0.95 & 1.03\\
		RNN-\modelname & 0.26 & 0.40 & 0.67 & 0.78 & 0.21 & 0.34 & 0.55 & 0.69 & 0.26 & 0.48 & 0.96 & 0.94 & 0.30 & 0.66 & 0.95 & 1.05\\
		\hline
		
	\end{tabular}
	}
	
	\caption{\textbf{H3.6M results} for the commonly used \textit{walking}, \textit{eating}, \textit{smoking}, and \textit{discussion} activities across different prediction horizons. 
	Values correspond to the Euler angle metric measured \textit{at} the given time. ``Seq2seq-sampling'' and ``Seq2seq-sampling-sup'' models correspond to ``Residual unsup. (MA)'' and ``Residual sup. (MA)'' models in \cite{Martinez2017Motion}, respectively. Note the relative performance improvement for each base model when augmented with our \modelnamelayer.}
	\label{tab:h36m}
\end{table*}

\subsection{Models} 
The modular nature of our \modelnamelayer~allows for flexible deployment with a diverse set of base models. In our experiments, we test the layer with the following three representative architectures proposed in the literature. To ease experimentation with \modelname~and other base architectures, we make all code and pre-trained models available at {\small{\url{https://ait.ethz.ch/projects/2019/spl}}}.

\noindent \textbf{Seq2seq } is a model proposed by Martinez \etal \cite{Martinez2017Motion}, consisting of a single layer of GRU cells. It contains a residual connection between the inputs and predictions. Input poses are represented as exponential maps.

\noindent \textbf{QuaterNet } uses a quaternion representation instead \cite{Pavllo2019Arxiv, Pavllo2018}. The model augments RNNs with quaternion based normalization and regularization operations. Similarly, the residual connection from inputs to outputs is implemented via the quaternion product. In our experiments, we replace the final linear output layer with our \modelnamelayer~and keep the remaining setup intact.

\noindent \textbf{RNN } uses a single layer recurrent network to calculate the context $\myvector{h}_t$, which we feed to our \modelnamelayer. In contrast to the Seq2seq and QuaterNet settings, we represent poses via rotation matrices. To account for the error accumulation problem at test time \cite{Fragkiadaki2015ERD, Ghosh2017, Jain2016}, we apply dropout directly on the inputs. This architecture is similar to the ERD \cite{Fragkiadaki2015ERD} but is additionally augmented with the residual connection of \cite{Martinez2017Motion}.

In the \modelnamelayer, each joint is modelled with only one small hidden layer (64 or 128 units) followed by a ReLU activation and a linear projection to the joint prediction $\jointhat{k} \in \real^M$. We experiment with different hierarchical configurations in \modelname~(cf. \secref{sec:ablation}) where following the true kinematic chain performed best. Some models benefit from inputting all parent joints in the kinematic chain compared to using only the immediate parent.
Note that we changed existing Seq2seq and QuaterNet models only as much as required to integrate them with \modelname. To ensure a fair comparison we fine-tune hyper-parameters like learning rate, batch size and hidden layer units. See appendix \secref{sec:app_arch_details} for details.

    \section{Evaluation on H3.6M Dataset}\label{sec:h36m}
In our first set of comparisons we baseline the proposed \modelnamelayer~on the H3.6M dataset using the Euler angle metric as is common practice in the literature.

\subsection{Metrics}
\label{sec:metrics-hmm}

\paragraph*{Euler angles}
Let $ \myvector{w} = \theta \myvector{a}$ denote a rotation of angle $\theta$ around the unit axis $\myvector{a} \in \mathbb{R}^3$. $\myvector{w}$ is the angle-axis (or exponential map) representation of a single joint angle. The Euler angles are extracted from $\myvector{w}$ by first converting it into a rotation matrix $\myvector{R} = \exp(\myvector{w})$ using Rodrigues' formula and then computing the angles $\boldsymbol{\alpha} = (\alpha_x, \alpha_y, \alpha_z)$ following \cite{EulerAngles}. This assumes that $\myvector{R}$ follows the \textit{z-y-x} order. Furthermore, as noted by \cite{EulerAngles}, there exist always two solutions for $\boldsymbol{\alpha}$, from which \cite{Jain2016} picks the one that leads to the least amount of rotation. The Euler angle metric for time step $t$ is then
\begin{equation}
L_{eul}(t) = \frac{1}{|\mathcal{X}_{test}|}\sum_{\pose \in \mathcal{X}_{test}} \sqrt{ \sum_k (\boldsymbol{\alpha}_t^{(k)} - \boldsymbol{\hat{\alpha}}_t^{(k)})^2 }
\label{eq:euler}
\end{equation}
where $\boldsymbol{\alpha}_t^{(k)}$ are the predicted Euler angles of joint $k$ at time $t$. $\mathcal{X}_{test}$ is defined by \cite{Jain2016} and comprises of 120 sequences.

\subsection{Results}

\tabref{tab:h36m} summarizes the relative performances of models with and without the \modelnamelayer~on the H3.6M dataset and compares them to the state of the art. The publicly available Seq2seq \cite{Martinez2017Motion} and QuaterNet \cite{Pavllo2018} models are augmented with our \modelnamelayer, but we otherwise follow the original training and evaluation protocols of the respective baseline model. 

Using the \modelnamelayer~improves the Seq2seq performance significantly and achieves state-of-the-art performance in the \textit{walking} category. Similarly, \modelname~yields the best performance with QuaterNet in short-term \textit{smoking} and \textit{discussion} motions and marginally outperforms the vanilla QuaterNet in most categories or is comparative to it. While our \modelnamelayer~also boosts the performance of the RNN model in \textit{walking}, \textit{eating} and \textit{smoking} motion categories, performance remains similar for \textit{discussion}. 

We follow the same evaluation setting as in previous work for direct comparability. It is noteworthy to mention that the evaluation metrics reported on H3.6M exhibit high variance due to the small number of test samples \cite{Pavllo2019Arxiv} and low errors do not always correspond to good qualitative results \cite{Martinez2017Motion}. 
    \section{AMASS: A New Benchmark}\label{sec:amass}
In this section we evaluate the baseline methods and our \modelnamelayer~on the large-scale AMASS dataset, detailed in \secref{sec:dataset}. The diversity and large amount of motion samples in AMASS increase both the task's complexity and the reliability of results due to a larger test set. In addition to proposing a new evaluation setting for motion modelling we suggest usage of a more versatile set of metrics for the task.

\begin{table*}[t!]
	\renewcommand{\arraystretch}{1.1}
	\setlength\tabcolsep{4.3pt}%
	\small{
	\begin{tabular} {l  r r r r | r r r r | r r r r | r r r r }
		& \multicolumn{4}{c}{Euler} & \multicolumn{4}{c}{Joint Angle} & \multicolumn{4}{c}{Positional} & \multicolumn{4}{c}{PCK (AUC)} \\       
		milliseconds & 100 & 200 & 300 & 400 & 100 & 200 & 300 & 400 & 100 & 200 & 300 & 400 & 100 & 200 & 300 & 400 \\
		\hline
		Zero-Velocity \cite{Martinez2017Motion} & 1.91 & 5.93 & 11.36 & 17.78 & 0.37 & 1.22 & 2.44 & 3.94 & 0.14 & 0.48 & 0.96 & 1.54 & 0.86 & 0.83 & 0.84 & 0.82\\
		\hline
		Seq2seq \cite{Martinez2017Motion}* & 1.52 & 5.14 & 10.66 & 17.84 & 0.27 & 0.99 & 2.19 & 3.85 & 0.11 & 0.39 & 0.87 & 1.53 & 0.91 & 0.86 & 0.86 & 0.82\\
		Seq2seq-PJL & 1.46 & 5.28 & 11.46 & 19.78 & 0.24 & 0.95 & 2.16 & 3.87 & 0.09 & 0.35 & 0.80 & 1.41 & 0.91 & 0.87 & 0.87 & 0.83 \\
		Seq2seq-\modelname & 1.57 & 5.00 & 10.01 & 16.43 & 0.27 & 0.94 & 2.01 & 3.45 & 0.10 & 0.36 & 0.79 & 1.36 & 0.91 & 0.87 & 0.87 & 0.84\\
		\hline
		Seq2seq-sampling \cite{Martinez2017Motion}* & 2.01 & 5.99 & 11.22 & 17.33 & 0.37 & 1.17 & 2.27 & 3.59 & 0.14 & 0.45 & 0.88 & 1.39 & 0.86 & 0.84 & 0.85 & 0.83\\
		Seq2seq-sampling-PJL & 1.71 & 5.15 & 9.71 & 15.15 & 0.32 & 1.00 & 1.97 & 3.14 & 0.12 & 0.39 & 0.77 & 1.23 & 0.88 & 0.86 & 0.87 & 0.85\\
		Seq2seq-sampling-\modelname & 1.71 & 5.13 & 9.60 & 14.86 & 0.31 & 0.97 & 1.91 & 3.04 & 0.12 & 0.38 & 0.74 & 1.18 & 0.89 & 0.86 & 0.88 & 0.85\\
		\hline
		Seq2seq-dropout & 1.54 & 4.98 & 9.94 & 16.13 & 0.27 & 0.95 & 2.00 & 3.39 & 0.10 & 0.37 & 0.79 & 1.34 & 0.91 & 0.87 & 0.87 & 0.84\\
		Seq2seq-dropout-PJL & 1.26 & 4.41 & 9.24 & 15.46 & 0.23 & 0.84 & 1.82 & 3.13 & 0.09 & 0.33 & 0.71 & 1.21 & 0.92 & 0.88 & 0.88 & 0.85\\
		Seq2seq-dropout-\modelname & \textbf{1.26} & 4.26 & 8.67 & 14.23 & 0.23 & 0.81 & 1.74 & 2.96 & 0.09 & 0.32 & 0.68 & 1.16 & 0.92 & 0.89 & 0.89 & 0.86\\
		\hline
		QuaterNet \cite{Pavllo2018}* & 1.49 & 4.70 & 9.16 & 14.54 & 0.26 & 0.89 & 1.83 & 3.00 & 0.10 & 0.34 & 0.71 & 1.18 & 0.90 & 0.87 & 0.88 & 0.85\\
		QuaterNet-\modelname & 1.34 & 4.25 & 8.39 & 13.43 & 0.25 & 0.83 & 1.71 & 2.83 & 0.09 & 0.32 & 0.67 & 1.10 & 0.91 & 0.88 & 0.89 & 0.86\\
		\hline
		RNN & 1.69 & 5.23 & 10.18 & 16.29 & 0.31 & 1.05 & 2.17 & 3.62 & 0.12 & 0.41 & 0.85 & 1.43 & 0.89 & 0.85 & 0.86 & 0.83\\
		RNN-\modelname & 1.33 & \textbf{4.13} & \textbf{8.03} & \textbf{12.84} & \textbf{0.22} & \textbf{0.73} & \textbf{1.51} & \textbf{2.51} & \textbf{0.08} & \textbf{0.28} & \textbf{0.57} & \textbf{0.96} & \textbf{0.93} & \textbf{0.90} & \textbf{0.90} & \textbf{0.88}\\
		\hline
		
	\end{tabular}
	}
	
	\caption{\textbf{AMASS results} of the base models with and without the proposed \modelnamelayer. We report normalized area-under-the-curve (AUC) for PCK values (higher is better, maximum is $1$). For the remaining metrics, lower is better. ``Seq2seq'' and ``Seq2seq-dropout'' are trained by using ground-truth inputs. "-dropout" applies $0.1$ dropout on the inputs. ``*'' indicates our evaluation of this model. "-PJL" stands for our proposed \textit{per-joint loss} on the vanilla model, showing a significant improvement already. Note that models with \modelname~perform better except on short-term predictions for ``Seq2seq'' model.}
	\label{tab:amass}
	\vspace{-0.2cm}
\end{table*}

\subsection{Metrics}\label{sec:metrics-amass}
So far, motion prediction has been benchmarked on H3.6M using the Euclidean distance between target and predicted Euler angles \cite{Jain2016, Martinez2017Motion, Pavllo2018, Wang2018Adversarial}. Numbers are usually reported per action at certain time steps averaged over 8 samples \cite{Jain2016}. Unfortunately, Euler angles have twelve different conventions (not counting the fact that each of these can be defined using intrinsic or extrinsic rotations), which makes the practical implementation of this metric error-prone. 

For a more precise analysis we introduce additional metrics from related pose estimation areas \cite{Spurr2018, SIP, Zimmermann2017}. In order to increase the robustness we furthermore suggest to
\begin{inparaenum}[i)]
\item sum \textit{until} time step $t$ rather than report the metric \textit{at} time step $t$,
\item use more test samples covering a larger portion of the test data set and
\item evaluate the models with complementary metrics.
\end{inparaenum}
Note that we do not train the models on these metrics; they only serve as evaluation criteria at test time.

\paragraph*{Joint angle difference}
To circumvent the potential source of error in the Euler angle metric, we propose using another angle-based metric following \cite{DIP, SIP}. This metric computes the angle of the rotation required to align the predicted joint with the target joint. Unlike $L_{eul}$, this metric is independent of how rotations are parameterized. It is furthermore similar to the geodesic loss proposed by \cite{Wang2018Adversarial}. Let $\myvector{\hat{R}}$ be the predicted joint angle for a given joint, parameterized as a rotation matrix, and the respective target rotation $\myvector{R}$. The difference in rotation can be computed as $\myvector{\tilde{R}} = \myvector{\hat{R}}\myvector{R}^T$, from which we construct the metric at time step $t$ as follows:
\begin{equation}
    L_{angle}(t) = \frac{1}{|\mathcal{X}_{test}|}\sum_{\pose \in \mathcal{X}_{test}} \frac{1}{K}\sum_k \norm{ \log\left(\myvector{\tilde{R}}_t^{(k)}\right) }_2 
    \label{eq:angular}
\end{equation}
where $\myvector{\tilde{R}}_t^{(k)}$ is the rotation matrix of joint $k$ at time $t$. In contrast to $L_{eul}$ we compute the loss on global joint angles by first unrolling the kinematic chain before computing $L_{angle}$.

\paragraph*{Positional}
Following Pavllo \etal's \cite{Pavllo2018} suggestion, we introduce a positional metric. This metric simply performs forward kinematics on $\pose$ and $\posehat$ to obtain 3D joint positions $\posepos$ and $\poseposhat$, respectively. It then computes the Euclidean distance per joint. We normalize the skeleton bones such that the right thigh bone has unit length.
\begin{equation}
    L_{pos}(t) = \frac{1}{|\mathcal{X}_{test}|}\sum_{\pose \in \mathcal{X}_{test}} \frac{1}{K}\sum_k \norm{\jointpos{k} - \jointposhat{k} }_2
\end{equation}

\paragraph*{PCK}
In cases where large errors occur, the value of $L_{pos}$ can be misleading. Hence, following the 3D (hand) pose estimation literature \cite{Iqbal2018PCK, Mueller2018PCK, Spurr2018, Zimmermann2017}, we introduce PCK by computing the percentage of predicted joints lying within a spherical threshold $\rho$ around the target joint position, i.e.
\begin{align}
    \textit{PCK}(\pose, \posehat, \rho) &= \frac{1}{K}\sum_k \mathbb{I} \left[ \norm{ \jointpos{k} - \jointposhat{k} }_2 \leq \rho \right] \nonumber \\
    L_{pck}(t, \rho) &= \frac{1}{|\mathcal{X}_{test}|}\sum_{\pose \in \mathcal{X}_{test}} \textit{PCK}(\pose, \posehat, \rho)
    \label{eq:pck}
\end{align}
where $\mathbb{I}[\cdot]$ returns $1$ if its input is true, and $0$ otherwise. Note that for PCK we do not sum, but average, until time step $t$.

\subsection{Results}
\tabref{tab:amass} summarizes the performance of the three model variants, each with and without the \modelnamelayer.
We trained the base models with minimal modifications, i.e. design, training objective and regularizations are kept intact. We use angle-axis, quaternion and rotation matrix representations for Seq2seq, QuaterNet, and RNN models, respectively. To make a fair comparison, we run hyper-parameter search on the batch size, cell type, learning rate and hidden layer size.

\begin{figure}[b]
	\centering
	\includegraphics[trim=0 290 290 0, clip,width=1.0\linewidth]{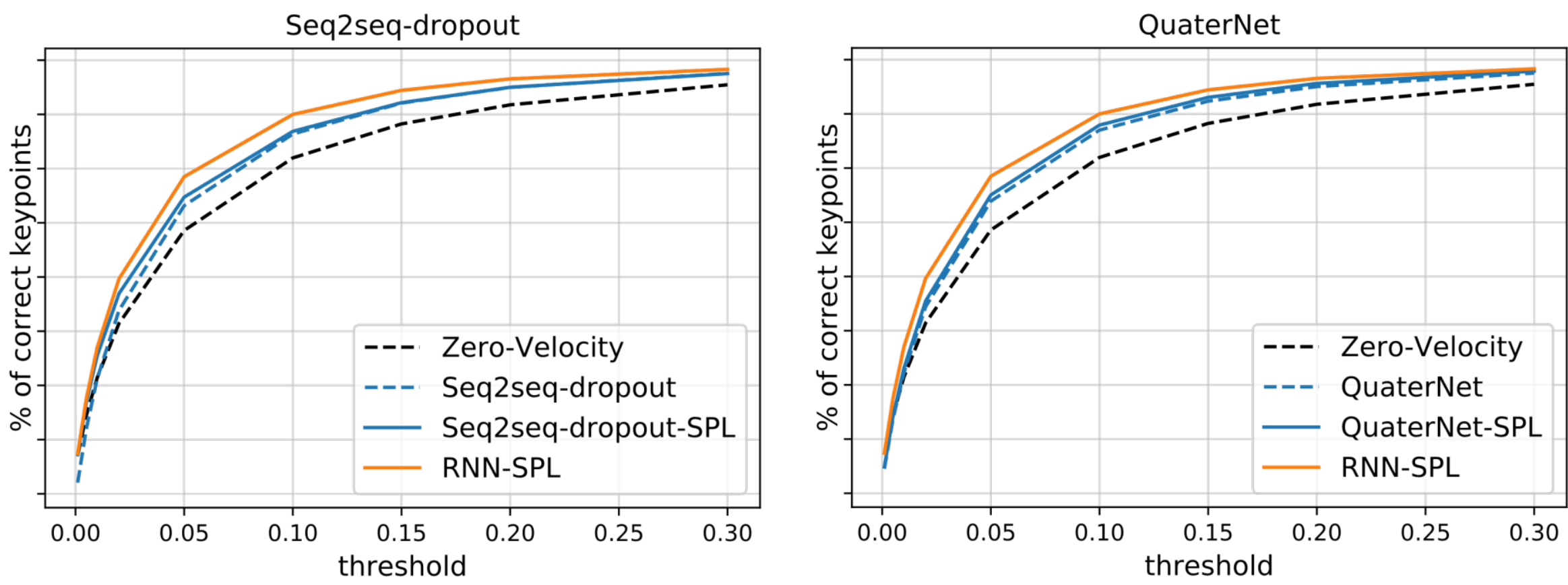}
	\caption{\textbf{PCK curves} of the best Seq2seq variant and QuaterNet with and without \modelname~on AMASS for $400$ ms predictions. More results are shown in appendix \secref{sec:app_pck}.}
	\label{fig:pck}
\end{figure}

\begin{figure*}
	\centering
	\includegraphics[trim=0 10 0 0, clip, width=2.0\columnwidth]{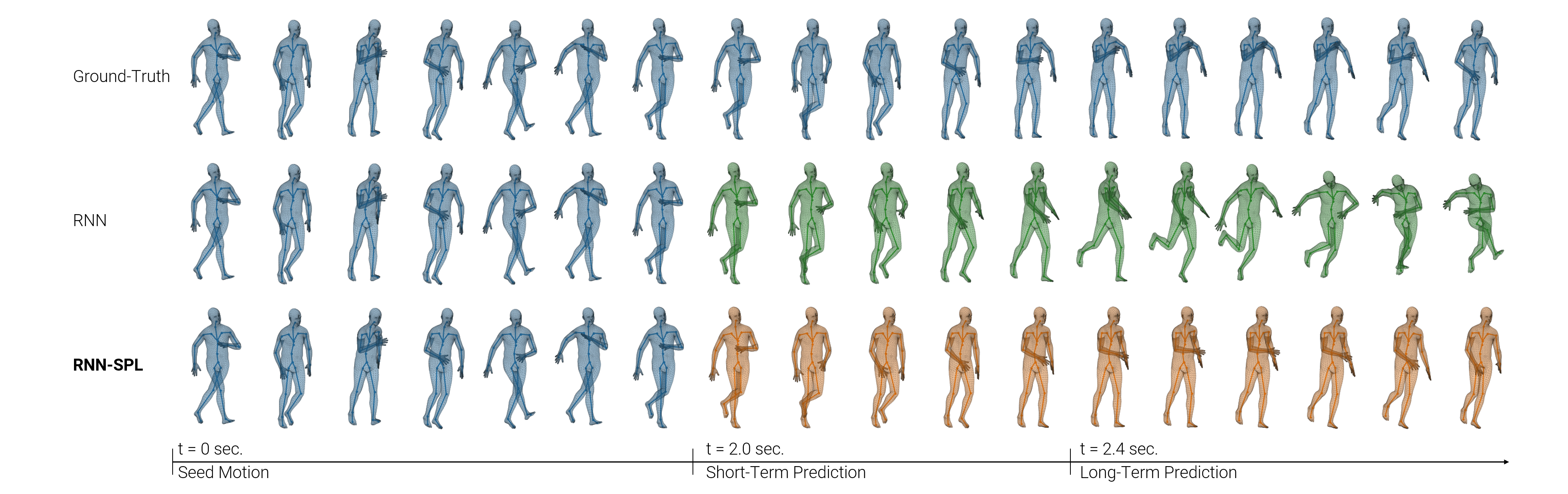}
	\caption{\textbf{Qualitative Comparison on AMASS.} We use a $2$-second seed sequence and predict the next $1$ second (60 frames). The last pose of the seed and the first pose of the prediction sequences are consecutive frames. Note that there is no transition problem. \textit{Top:} Ground-truth sequence. \textit{Middle:} Output of the vanilla RNN which quickly deteriorates. \textit{Bottom:} The same RNN model augmented with our \modelnamelayer. It produces accurate short-term predictions as well as natural long-term motion.}
	\label{fig:qualitative}
	\vspace{-0.2cm}
\end{figure*}

Unlike on H3.6M, LSTM cells consistently outperform GRUs on AMASS for the Seq2seq and RNN models. Different from \cite{Martinez2017Motion}, we also train the Seq2seq model by applying dropout on the inputs similar to our RNN architecture. QuaterNet gives its best performance with GRU cells while some fine-tuning for the teacher forcing ratio is necessary.

In all settings, the Seq2seq models fail to give competitive performance on this large-scale task and are sometimes outperformed by the zero-velocity baseline proposed by Martinez \etal \cite{Martinez2017Motion}.
QuaterNet shows a strong performance and is in fact the closest vanilla model to the \modelname~variants. However, our \modelnamelayer~still improves the QuaterNet results further. The contribution of the \modelnamelayer~is best observable on the RNN model. With the help of a larger dataset, the proposed RNN-\modelname~achieves the best results under different metrics and prediction horizons. \figref{fig:pck} compares two baseline methods for $400$ millisecond predictions with their corresponding \modelname~extension for different choices of the threshold $\rho$. The RNN-\modelname~consistently outperforms other methods. More results are shown in the appendix \secref{sec:app_pck}.

Please also note the complementary effect of the proposed metrics in \tabref{tab:amass}. The Seq2seq-dropout-\modelname~model at $100$ ms shows a significant improvement ($1.26$) w.r.t. the Euler angle metric, and in fact achieves the best result across all models. However, this is no longer the case when we look at the proposed metrics. The model performs marginally worse than the best performing model, RNN-\modelname, in these metrics. The joints closer to the root of the kinematic chain have a much larger impact on the overall pose since wrong rotations propagate to all the child joints on the chain. This effect might be ignored when only local rotations are considered, which is the case for $L_{eul}$. $L_{angle}$ and $L_{pos}$ account for this by first unrolling the kinematic chain.

In line with \cite{Pavllo2018, Wang2018Adversarial}, we report that the residual connection from \cite{Martinez2017Motion} is very effective for short-term predictions. All models we trained performed better with the residual connection irrespective of the dataset or pose representation. 

    \subsection{Ablation Study}\label{sec:ablation}
To study the \modelname~in more depth we conduct an ablation study presented in \tabref{tab:ablation_small}. We observe that the main performance boost is achieved by the decomposition of the output layer and the per-joint loss in Eq. \eqref{eq:jointwise_loss}. While the per-joint-loss alone (i.e., without \modelname) is not beneficial on H3.6M, on AMASS its application alone already helps (\textit{RNN-PJL}). It is also effective on Seq2seq models with noisy inputs, but the performance degrades on vanilla Seq2seq model. In longer-term predictions, SP-layer shows a significant contribution (see \tabref{tab:amass}).
Assuming independent joints without modelling any hierarchy (\textit{RNN-SPL-indep.}) improves the results further. Introducing hierarchy into the prediction layer either in reverse or random order performs often similar or better. However, introducing the spatial dependencies according to the kinematic chain (\textit{RNN-SPL}) yields the best results with the exception of the positional metric. 

\begin{table}[t]
    \renewcommand{\arraystretch}{1.1}
    \setlength\tabcolsep{4.8pt}
    \small{
    \begin{tabular}{l c c c || c}
             & \multicolumn{3}{c||}{AMASS} & H3.6M \\
             & Euler & Joint Angle & Pos. & Walking \\
    \hline
    RNN & 16.44 & 3.570 & 1.396 & 0.900\\
    RNN-PJL & 13.13 & 2.573 & 0.986 & 0.950\\
     \hline
    RNN-SPL-indep. & 12.96 & 2.552 & 0.982 & 0.836\\
    RNN-SPL-random & 12.98 & 2.547 & 0.980  & 0.863\\
    RNN-SPL-reverse & 13.03 & 2.543 & \textbf{0.973} & 0.849\\
     \hline
    RNN-SPL & \textbf{12.85} & \textbf{2.533} & 0.975 & \textbf{0.772}\\
     \hline
    \end{tabular}
    }
    \caption{\textbf{Ablation study} on AMASS and H3.6M (\textit{walking}) for $400$ ms predictions. Each entry is an average over 5 randomly initialized training runs. Please refer to \secref{sec:ablation} for detailed explanations and the appendix for more results. 
    }
    \label{tab:ablation_small}
    \vspace{-0.5cm}
\end{table}

    \section{Conclusion}
We introduce prior knowledge about the human skeletal structure into a neural network by means of a structured prediction layer (\modelname). The \modelnamelayer~explicitly decomposes the pose into individual joints and can be interfaced with a variety of baseline architectures.
We furthermore introduce AMASS, a large-scale motion dataset, and several metrics to the task of motion prediction. On AMASS, we empirically show that for any baseline model, any metric, and any input representation, it is better to use the proposed \modelnamelayer. The simple RNN model augmented with the \modelnamelayer~achieved state-of-the-art performance on the new AMASS benchmark.
    
    {

    \paragraph*{Acknowledgements} We thank the reviewers for their insightful comments and Martin Blapp for fruitful discussions. This project has received funding from the European Research Council (ERC) under the European Union’s Horizon 2020 research and innovation programme grant agreement No 717054. We thank the NVIDIA Corporation for the donation of GPUs used in this work.
    }
    
    {\small
    \bibliographystyle{ieee_fullname}
    \bibliography{egbib}
    }
    
    \clearpage
    
    \ifsupplementary
        \section{Appendix}
We provide architecture details in \secref{sec:app_arch_details}, results on long-term predictions in \secref{sec:app_longterm}, PCK plots in \secref{sec:app_pck}, and more detailed ablation studies in \secref{sec:app_ablation}.

\subsection{Architecture Details}
\label{sec:app_arch_details}
The RNN and Seq2seq models are implemented in Tensorflow \cite{tensorflow}.  For the QuaterNet-\modelname~model we extend the publicly available source code in Pytorch \cite{pytorch}. Our aim is to make a minimum amount of modifications to the baseline Seq2seq \cite{Martinez2017Motion} and QuaterNet \cite{Pavllo2018} models. In order to get the best performance on the new AMASS dataset, we fine-tune the hyper-parameters including batch size, learning rate, learning rate decay, cell type and number of cell units, dropout rate, hidden output layer size and teacher forcing ratio decay for QuaterNet.

\figref{fig:model-overview} provides an overview over these models. The \modelnamelayer~replaces the standard dense layers, which normally use the context representation $\myvector{h}_t$, i.e., GRU or LSTM state until time-step $t$, to make the pose vector prediction $\posehat$. The \modelname~component follows the kinematic chain and uses the following network for every joint:
\begin{equation*}
Linear(H)-ReLU-Linear(M)~,
\end{equation*}
where the hidden layer size per joint $H$ is either $64$ or $128$ and the joint size $M$ is $3$, $4$, or $9$ for exponential map, quaternion, or rotation matrix pose representation, respectively (see Tab. \ref{tab:architecture}). Similar to the H3.6M setup \cite{Jain2016, Martinez2017Motion} we use a $2$-second seed $\poset{1:t-1}$ and $400$-milisecond target sequences $\poset{t:T}$. The sequence $\myvector{x}_{t:T}$ corresponds to the target predictions.

\begin{table}[b]
	\renewcommand{\arraystretch}{1.1}
	\setlength\tabcolsep{4.0pt}%
	\small{\begin{tabular} {l  c c c | c c c }
		& \multicolumn{3}{c}{H3.6M} & \multicolumn{3}{c}{AMASS} \\       
		 & SPL & Units & Cell & SPL & Units & Cell \\
		\hline
		RNN-\modelname & sparse & 64 & GRU & dense & 64 & LSTM \\
		\hline
		Seq2seq-\modelname & sparse & 64 & GRU & dense & 64 & LSTM \\
		\hline
		QuaterNet-\modelname & sparse & 128 & GRU & sparse & 128 & GRU \\
		\hline
		
	\end{tabular}
	}
	\caption{\textbf{SPL configuration.} \textit{sparse} and \textit{dense} refer to making a joint prediction by feeding only the immediate parent or all parent joints in the kinematic chain, respectively. Models use a hidden layer of either $64$ or $128$ units per joint. GRU cell outperforms LSTM on H3.6M while LSTM is consistently better on AMASS dataset. The vanilla models use their original setting with the reported cell.}
	\label{tab:architecture}
\end{table}

We train the baseline Seq2seq \cite{Martinez2017Motion} and QuaterNet \cite{Pavllo2018} models by using the training objectives as proposed in the original papers. The \modelname~variants, however, implement these objectives by using our proposed joint-wise loss. After an epoch of training we evaluate the model on the validation split and apply early stopping with respect to the joint angle metric. Please note that the early stopping metric is different than the training objective for all models.

\begin{figure}[t]
	\centering
	\includegraphics[width=0.99\columnwidth, trim={0pt 0pt 0pt 0pt}]{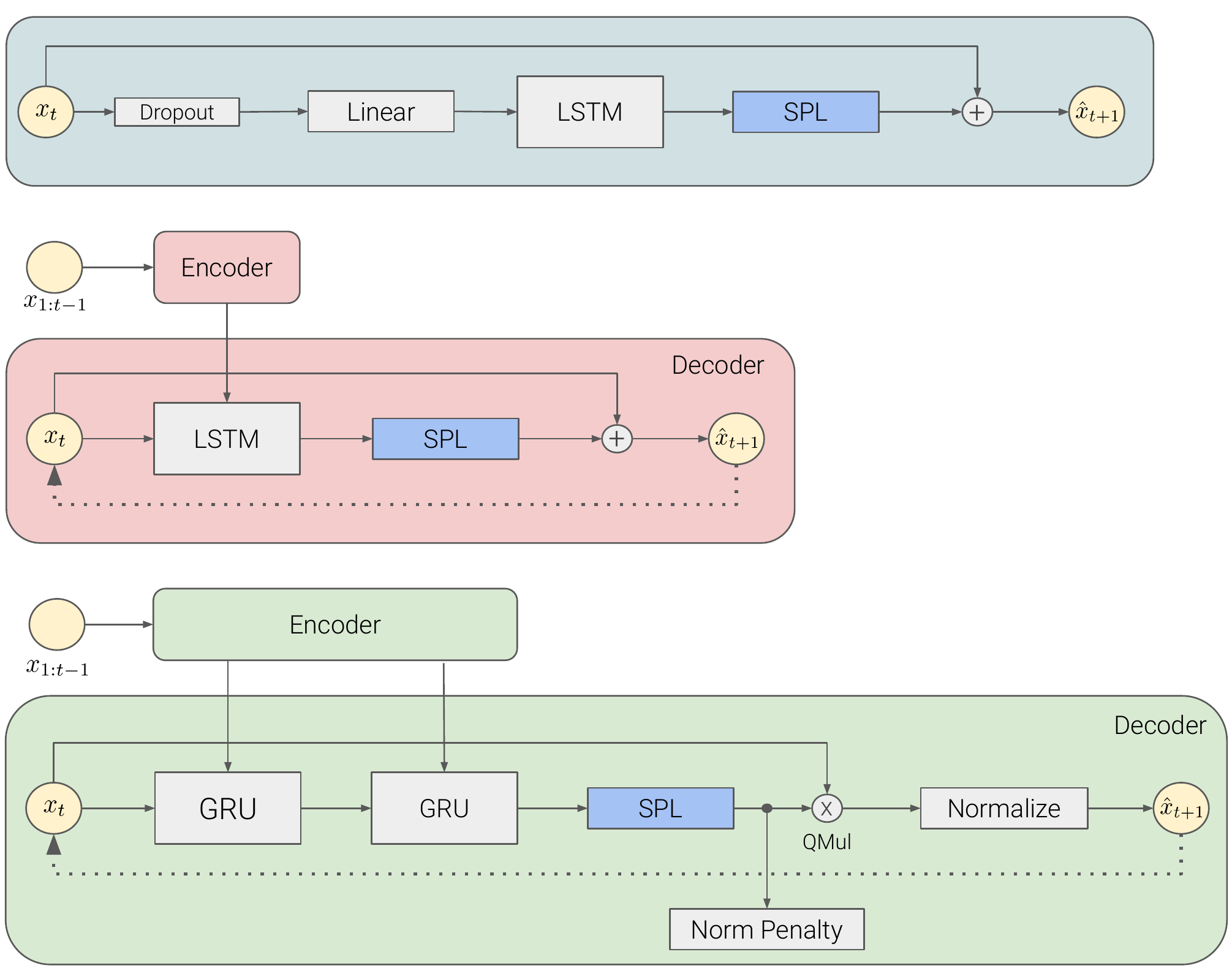}
	\caption{\textbf{Model overview}. \textit{Top:} RNN-\modelname~ \textit{Middle:} Seq2seq-\modelname, \textit{Bottom:} Quaternet-\modelname. Note that both Seq2seq and QuaterNet models follow sequence-to-sequence architecture where the encoder and decoder share the parameters. The $2$-second seed sequence $\poset{1:t-1}$ is first fed to the encoder network to calculate the hidden cell state which is later used to initialize the prediction into the future. The dashed lines from the prediction to the input correspond to the sampling based training. In other words, the predictions are fed back during training.}
	\label{fig:model-overview}
\end{figure}

\begin{table*}[t!]
\centering
	\renewcommand{\arraystretch}{1.1}
	\setlength\tabcolsep{4.5pt}%
	\begin{tabular} {l  r r r | r r r | r r r | r r r }
		& \multicolumn{3}{c}{Euler} & \multicolumn{3}{c}{Joint Angle} & \multicolumn{3}{c}{Positional} & \multicolumn{3}{c}{PCK (AUC)} \\       
		milliseconds & 600 & 800 & 1000 & 600 & 800 & 1000 & 600 & 800 & 1000 & 600 & 800 & 1000 \\
		\hline
		Zero-Velocity \cite{Martinez2017Motion} & 32.36 & 48.39 & 65.25 & 7.46 & 11.31 & 15.3 & 2.93 & 4.46 & 6.06 & 0.78 & 0.76 & 0.74\\
		\hline
		Seq2seq \cite{Martinez2017Motion}* & 36.39 & 60.07 & 88.72 & 8.39 & 14.36 & 21.61 & 3.38 & 5.82 & 8.81 & 0.75 & 0.71 & 0.67\\
		Seq2seq-PJL & 41.96 & 71.63 & 109.45 & 8.75 & 15.57 & 24.43 & 3.13 & 5.55 & 8.76 & 0.76 & 0.71 & 0.66\\
		Seq2seq-\modelname & 32.58 & 52.49 & 75.69 & 7.23 & 11.99 & 17.62 & 2.88 & 4.81 & 7.10 & 0.79 & 0.75 & 0.72\\
		\hline
		Seq2seq-sampling \cite{Martinez2017Motion}* & 31.37 & 47.37 & 64.72 & 6.72 & 10.31 & 14.23 & 2.61 & 4.03 & 5.58 & 0.79 & 0.77 & 0.75\\
		Seq2seq-sampling-PJL & 27.72 & 42.19 & 58.01 & 5.96 & 9.21 & 12.79 & 2.34 & 3.64 & 5.07 & 0.81 & 0.79 & 0.77\\
		Seq2seq-sampling-\modelname & 27.01 & 40.90 & 55.97 & 5.76 & 8.90 & 12.36 & 2.24 & 3.48 & 4.85 & 0.82 & 0.80 & 0.78\\
		\hline
		Seq2seq-dropout & 31.16 & 48.92 & 68.77 & 6.94 & 11.22 & 16.06 & 2.78 & 4.54 & 6.54 & 0.78 & 0.75 & 0.72\\
		Seq2seq-dropout-PJL & 31.20 & 50.62 & 73.09 & 6.59 & 10.93 & 15.98 & 2.53 & 4.18 & 6.09 & 0.80 & 0.76 & 0.73\\
		Seq2seq-dropout-\modelname & 28.02 & 44.95 & 64.23 & 6.15 & 10.11 & 14.67 & 2.42 & 4.00 & 5.84 & 0.81 & 0.78 & 0.75\\
		\hline
		QuaterNet \cite{Pavllo2018}* & 27.08 & 41.32 & 56.66 & 5.88 & 9.21 & 12.84 & 2.32 & 3.64 & 5.09 & 0.82 & 0.79 & 0.77\\
		QuaterNet-\modelname & 25.37 & 39.02 & 53.95 & 5.58 & 8.79 & 12.32 & 2.19 & 3.47 & 4.87 & 0.82 & 0.80 & 0.78\\
		\hline
		RNN & 31.19 & 48.84 & 68.64 & 7.33 & 11.87 & 17.09 & 2.93 & 4.79 & 6.96 & 0.78 & 0.74 & 0.71\\
		RNN-\modelname & \textbf{24.44} & \textbf{38.02} & \textbf{53.06} & \textbf{5.04} & \textbf{8.08} & \textbf{11.50} & \textbf{1.94} & \textbf{3.14} & \textbf{4.49} & \textbf{0.84} & \textbf{0.81} & \textbf{0.79}\\
		\hline
		
	\end{tabular}
	
	\caption{\textbf{Long-term AMASS results} of the base models with and without the proposed structured prediction layer (\modelname). For PCK we report the area-under-the-curve (AUC), which is upper-bounded by $1$ (higher is better). Euler, joint angle and positional losses are lower-bounded by $0$ (lower is better). "*" indicates our evaluation of the corresponding model on AMASS. "-dropout" stands for dropout applied directly on the inputs. "-PJL" stands for our proposed \textit{per-joint loss} on the vanilla model, showing a significant improvement already. All models use residual connections. Note that models with our proposed \modelnamelayer~always perform better.}
	\label{tab:amass_longer}
\end{table*}

\paragraph{RNN-\modelname} We use the rotation matrix pose representation with zero-mean unit-variance normalization, following teacher-forcing training. In other words, the model is trained by feeding the ground-truth pose $\pose$ to predict $\myvector{\hat{x}}_{t+1}$. The training objective is the proposed joint-wise loss with $l_2$-norm (see \secref{sec:per_joint_loss} in the paper) which is calculated over the entire seed $\poset{1:t-1}$ and target predictions $\myvector{\hat{x}}_{t:T}$.

We do not follow a sampling-based training scheme. In the absence of such a training regularization, the model overfits to the likelihood (i.e., ground-truth input samples) and hence performs poorly in the auto-regressive test setup. We find that a small amount of dropout with a rate of $0.1$ on the inputs makes the model robust against the exposure bias problem. 

The dropout is followed by a linear layer with $256$ units. We use a single LSTM cell with $1024$ units. The vanilla RNN model makes the predictions by using 
\begin{equation*}
Linear(960)-ReLU-Linear(N)~,
\end{equation*}
where $N = K \cdot M$. We also experimented with GRU units instead of LSTM cells, but experimentally found that LSTMs consistently outperformed GRUs. Finally, we use the Adam optimizer \cite{kingma2014adam} with its default parameters. The learning rate is initialized with $1e^{-3}$ and exponentially decayed with a rate of $0.98$ at every $1000$ decay steps.

\paragraph{Seq2seq-\modelname} As proposed by Martinez \etal \cite{Martinez2017Motion} we use the exponential map pose representation with zero-mean unit-variance normalization. The model consists of encoder and decoder components where the parameters are shared. The seed sequence $\poset{1:t-1}$ is first fed to the encoder network to calculate the hidden cell state which is later used by the decoder to initialize the prediction into the future (i.e., $\myvector{\hat{x}}_{t:T}$). Similarly, the training objective is calculated between the ground-truth targets $\poset{t:T}$ and the predictions $\myvector{\hat{x}}_{t:T}$. We use the proposed joint-wise loss with $l_2$-norm.

In our AMASS experiments, we find that a single LSTM cell with $1024$ units performs better than a single GRU cell. In the training of the Seq2seq-sampling model, the decoder prediction is fed back to the model \cite{Martinez2017Motion}. The other two variants, Seq2seq-dropout (with a dropout rate of $0.1$) and Seq2seq (see Tab. 2 in the paper), are trained with ground-truth inputs similar to the RNN models. Similarly, the vanilla Seq2seq model has a hidden output layer of size $960$ on AMASS dataset.

We use the Adam optimizer with its default parameters. The learning rate is initialized with $1e^{-3}$ and exponentially decayed with a rate of $0.95$ at every $1000$ decay steps.

\paragraph{QuaterNet-\modelname} We use the quaternion pose representation without any further normalization on the data \cite{Pavllo2018}. The data is pre-processed following Pavllo \etal's suggestions to avoid mixing antipodal representations within a given sequence. QuaterNet also follows the sequence-to-sequence architecture where the seed sequence is used to initialize the cell states. As in the vanilla model, the training objective is based on the Euler angle pose representation. More specifically, the predictions in quaternion representation are converted to Euler angles to calculate the training objective.

The model consists of two stacked GRU cells with $1000$ units each. In contrast to the RNN and Seq2seq models, the residual velocity is implemented by using quaternion multiplication. Moreover, the QuaterNet model applies a normalization penalty and explicitly normalizes the predictions in order to enforce valid rotations. As proposed by Pavllo \etal \cite{Pavllo2018}, we exponentially decay the teacher-forcing ratio with a rate of $0.98$. The teacher-forcing ratio determines the probability of using ground-truth poses during training. Over time this value gets closer to $0$ and hence increases the probability of using the model predictions rather than the ground-truth poses. Similar to the vanilla RNN and Seq2seq models, a hidden output layer of size $960$ performed better on AMASS dataset.

Finally, the model is trained by using the Adam optimizer with its default parameters. The learning rate is initialized with $1e^{-3}$ and exponentially decayed with a rate of $0.96$ after every training epoch.

\begin{figure*}[t]
	\centering
	\includegraphics[trim=0 10 0 0, clip, width=2.0\columnwidth]{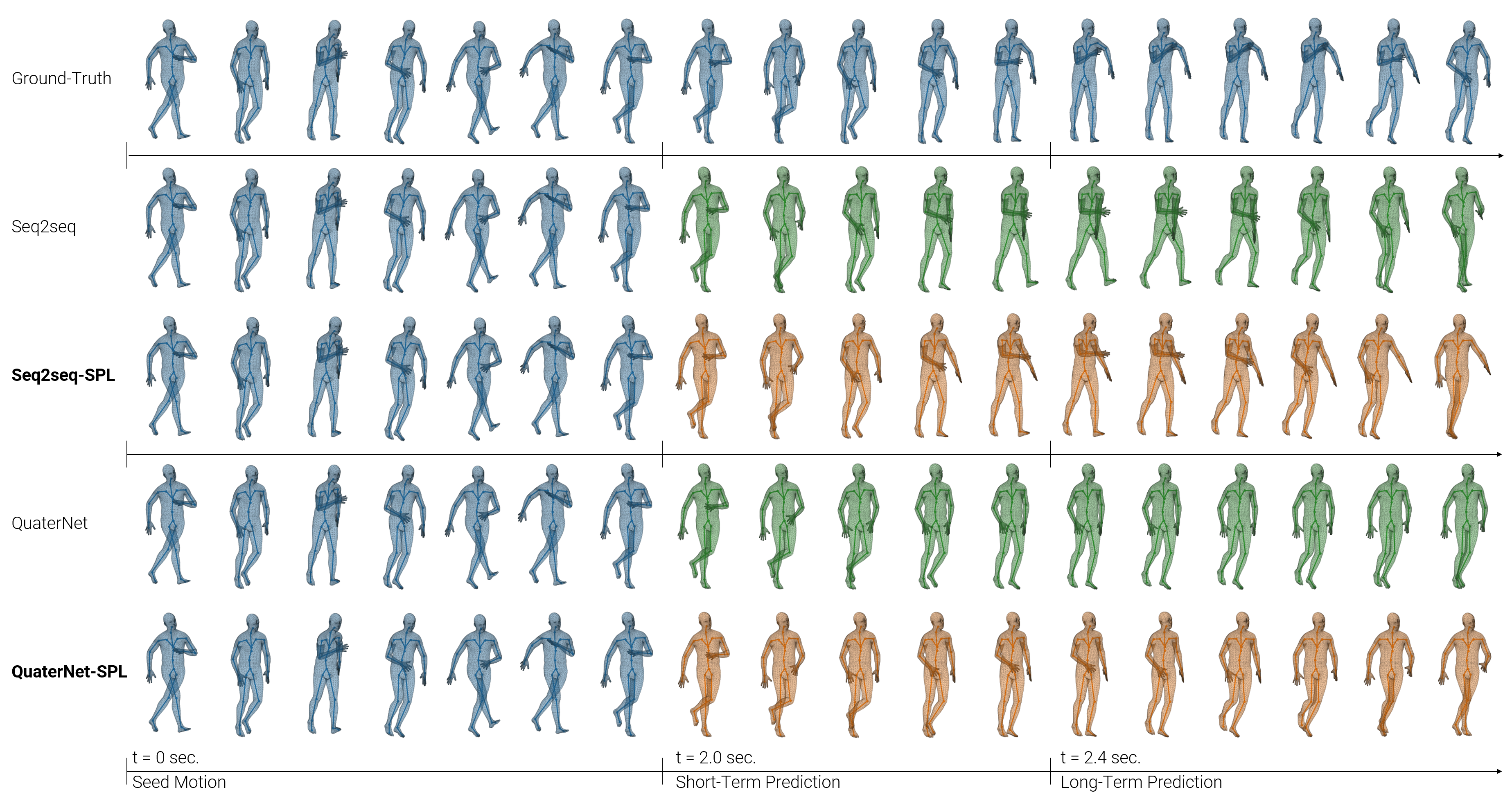}
	\caption{\textbf{Qualitative Comparison on AMASS.} We use a $2$-second seed sequence and predict the next $1$ second (60 frames). The last pose of the seed and the first pose of the prediction sequences are consecutive frames. In green (2nd and 4th row) are results from the vanilla versions of Seq2seq and QuaterNet, respectively. In orange (3rd and 5th row) are results when augmenting the vanilla model with our \modelnamelayer. Although the \modelname-variants shown here are still outperformed by the RNN-\modelname~shown in the main paper, they still show slight improvement over their non-\modelname~counterparts.}
	\label{fig:zoo}
\end{figure*}

\begin{table*}[h]
	\renewcommand{\arraystretch}{1.2}
	\setlength\tabcolsep{3.2pt}%
	\small{
	\begin{tabular} {l  r r r r | r r r r | r r r r | r r r r }
		& \multicolumn{4}{c}{Walking} & \multicolumn{4}{c}{Eating} & \multicolumn{4}{c}{Smoking} & \multicolumn{4}{c}{Discussion} \\       
		milliseconds & 80 & 160 & 320 & 400 & 80 & 160 & 320 & 400 & 80 & 160 & 320 & 400 & 80 & 160 & 320 & 400 \\
		\hline
		RNN-mean & 0.319 & 0.515 & 0.771 & 0.900 & 0.242 & 0.384 & 0.583 & 0.742 & 0.264 & 0.493 & 0.984 & 0.967 & 0.312 & 0.668 & 0.945 & 1.040\\
		\hline
		RNN-PJL & 0.324 & 0.534 & 0.816 & 0.950 & 0.233 & 0.391 & 0.616 & 0.776 & 0.258 & 0.483 & 0.961 & 0.932 & 0.312 & 0.675 & 0.969 & 1.067\\
		\hline
		RNN-\modelname-indep. & 0.288 & 0.453 & 0.720 & 0.836 & 0.228 & 0.366 & 0.575 & 0.736 & 0.258 & 0.482 & 0.947 & 0.916 & 0.313 & 0.676 & 0.962 & 1.064\\
		\hline
		RNN-\modelname-random & 0.298 & 0.473 & 0.758 & 0.863 & 0.227 & 0.354 & 0.578 & 0.717 & 0.263 & 0.490 & 0.956 & 0.925 & 0.311 & 0.677 & 0.975 & 1.079\\
		\hline
		RNN-\modelname-reverse & 0.302 & 0.483 & 0.725 & 0.849 & 0.225 & 0.344 & 0.557 & 0.721 & 0.264 & 0.494 & 0.96 & 0.929 & 0.312 & 0.679 & 0.960 & 1.050\\
		\hline
		RNN-\modelname & 0.264 & 0.413 & 0.669 & 0.772 & 0.205 & 0.326 & 0.559 & 0.721 & 0.260 & 0.486 & 0.958 & 0.930 & 0.307 & 0.667 & 0.950 & 1.049\\
		\hline
		
	\end{tabular}
	}
	
	\caption{\textbf{H3.6M ablation study.} Comparison of \modelname~with different joint configurations and the proposed per-joint loss on H3.6M. Each model entry corresponds to an average of several runs with different initialization. }
	\label{tab:ablation_h36m}
\end{table*}

\subsection{Long-term Prediction on AMASS}
\label{sec:app_longterm}

In \tabref{tab:amass_longer}, we report longer-term prediction results as an extension to the results provided in Tab. 2 in the main paper. Please note that all models are trained to predict $400$-ms. In fact, the Seq2seq and QuaterNet models have been proposed to solve short-term prediction tasks only. 

Consistent with the short-term prediction results shown in the main paper, our proposed \modelnamelayer~always improves the underlying model performance. While QuaterNet-\modelname~is competitive, RNN-\modelname~yields the best performance under different metrics.

In \figref{fig:zoo} we show more qualitative results for QuaterNet and Seq2seq when augmented with our \modelnamelayer. Please refer to the supplemental video for more qualitative results.

\subsection{PCK Plots}
\label{sec:app_pck}
We provide additional PCK plots for $100$, $200$, $300$ and $400$ ms prediction horizon in \figref{fig:pck_appendix}. Please note that shorter time horizons do not use the entire range of thresholds $\rho$ to avoid a saturation effect.

\begin{figure*}[b]
	\centering
	\includegraphics[trim=0 0 230 0,clip,width=2.0\columnwidth]{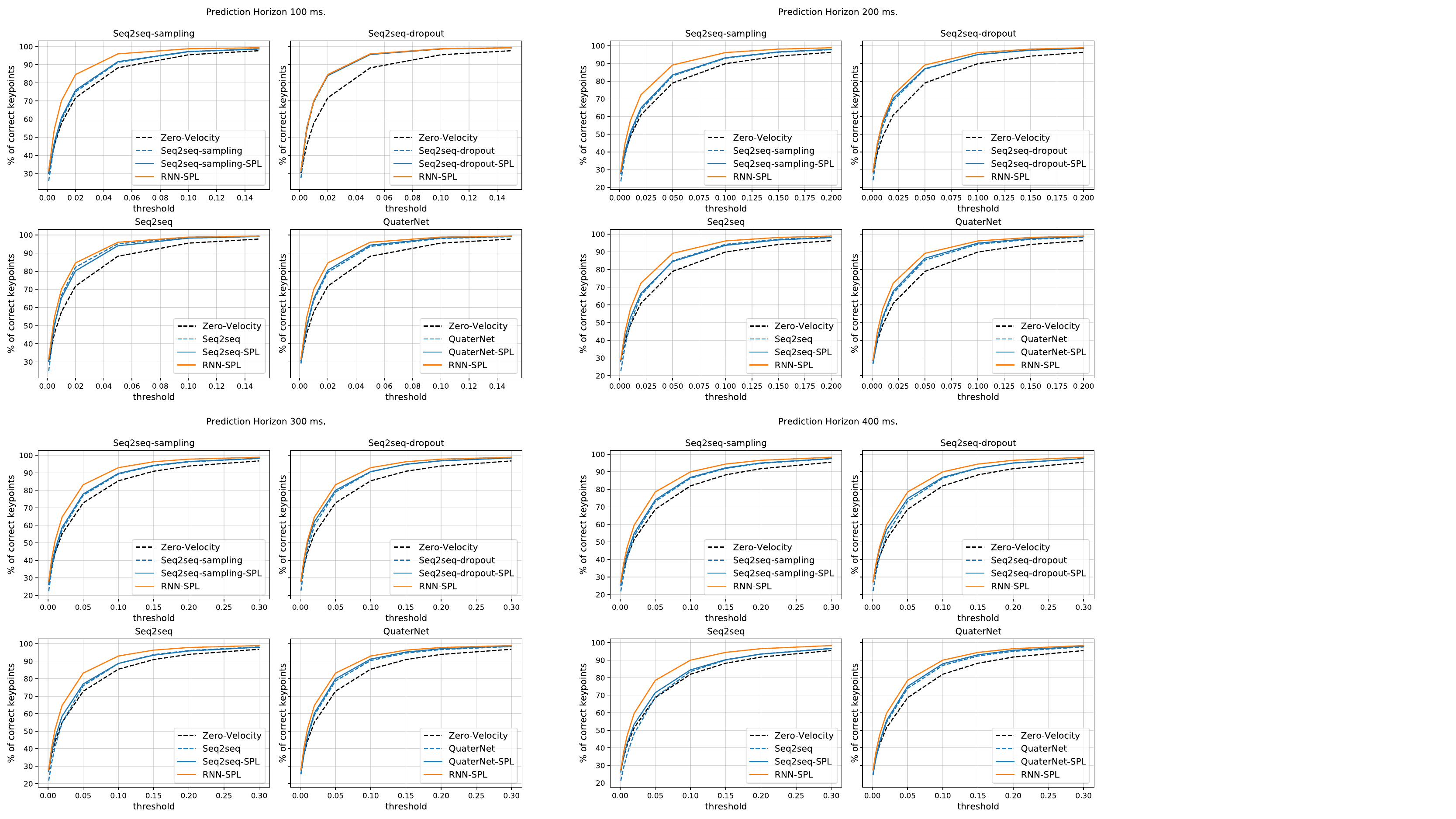}
	\caption{\textbf{PCK Curves} of models with and without our \modelnamelayer~(dashed lines) on AMASS for $100$, $200$, $300$, and $400$ milliseconds (\textit{top left} to \textit{bottom right}).}
	\label{fig:pck_appendix}
\end{figure*}

\subsection{Ablation Study}
\label{sec:app_ablation}
The full ablation study on H3.6M and AMASS is shown in \tabref{tab:ablation_h36m} and \ref{tab:amass_ablation}, respectively. For an explanation of each entry, please refer to the main text in \secref{sec:ablation}.

\begin{table*}[t]
	\renewcommand{\arraystretch}{1.2}
	\setlength\tabcolsep{3.2pt}%
	\small{
	\begin{tabular} {l  r r r r | r r r r | r r r r | r r r r }
		& \multicolumn{4}{c}{Euler} & \multicolumn{4}{c}{Joint Angle} & \multicolumn{4}{c}{Positional} & \multicolumn{4}{c}{PCK (AUC)} \\       
		milliseconds & 100 & 200 & 300 & 400 & 100 & 200 & 300 & 400 & 100 & 200 & 300 & 400 & 100 & 200 & 300 & 400 \\
		\hline
		RNN-mean  & 1.65 & 5.21 & 10.24 & 16.44 & 0.318 & 1.057 & 2.157 & 3.570 & 0.122 & 0.408 & 0.838 & 1.396 & 0.886 & 0.854 & 0.861 & 0.832\\
		\hline
		RNN-PJL & 1.33 & 4.15 & 8.16 & 13.13 & 0.230 & 0.758 & 1.550 & 2.573 & 0.086 & 0.287 & 0.590 & 0.986 & 0.923 & 0.897 & 0.901 & 0.877\\
		\hline
		RNN-\modelname-indep. & 1.30 & 4.08 & 8.04 & 12.96 & 0.228 & 0.750 & 1.537 & 2.552 & 0.085 & 0.283 & 0.587 & 0.982 & 0.924 & 0.897 & 0.901 & 0.878\\
		\hline
		RNN-\modelname-random & 1.31 & 4.09 & 8.03 & 12.98 & 0.228 & 0.749 & 1.533 & 2.547 & 0.086 & 0.284 & 0.586 & 0.980 & 0.924 & 0.897 & 0.901 & 0.878\\
		\hline
		RNN-\modelname-reverse & 1.31 & 4.10 & 8.08 & 13.03 & 0.229 & 0.749 & 1.532 & 2.543 & 0.086 & 0.282 & 0.582 & 0.973 & 0.924 & 0.897 & 0.902 & 0.878\\
		\hline
		RNN-\modelname & 1.29 & 4.04 & 7.95 & 12.85 & 0.227 & 0.744 & 1.525 & 2.533 & 0.085 & 0.282 & 0.582 & 0.975 & 0.924 & 0.898 & 0.902 & 0.878\\
		\hline
	\end{tabular}
	}
	\caption{\textbf{AMASS ablation study.} Comparison of \modelname~with different joint configurations and the proposed per-joint loss on AMASS. Each model entry corresponds to an average of several runs with different initialization.}
	\label{tab:amass_ablation}
\end{table*}

    \fi 
\else
    \title{Structured Prediction Helps 3D Human Motion Modelling \\ Supplementary Material}
    \maketitle
     
    \clearpage
        {\small
        \bibliographystyle{ieee}
        \bibliography{egbib}
        }
\fi 
\end{document}